\documentclass[letterpaper]{article} 
\usepackage{aaai2026}  
\usepackage{times}  
\usepackage{helvet}  
\usepackage{courier}  
\usepackage[hyphens]{url}  
\usepackage{graphicx} 
\usepackage{natbib}  
\usepackage{caption} 
\usepackage{algorithm}
\usepackage{algorithmic}
\usepackage{multirow}
\usepackage{amssymb}
\usepackage{amsmath}
\usepackage{booktabs}
\usepackage{siunitx}
\usepackage{makecell}
\usepackage{float}
\usepackage{tabularx}
\usepackage{array}
\usepackage{graphicx}
\usepackage{subcaption}
\usepackage{multicol}

\usepackage{newfloat}
\usepackage{listings}
\DeclareCaptionStyle{ruled}{labelfont=normalfont,labelsep=colon,strut=off} 
\floatstyle{ruled}
\newfloat{listing}{tb}{lst}{}
\floatname{listing}{Listing}

\nocopyright

\title{QualityFM: a Multimodal Physiological Signal Foundation Model with Self-Distillation for Signal Quality Challenges in Critically Ill Patients}

\author{
  Zongheng Guo\textsuperscript{1},
  Tao Chen\textsuperscript{2},
  Manuela Ferrario\textsuperscript{1}
  \\[2ex]
  \textsuperscript{1}{\normalfont Department of Electronics, Information and Bioengineering, Politecnico di Milano, Milan, Italy}\\
  \textsuperscript{2}{\normalfont State Key Laboratory of Industrial Control Technology, Zhejiang University, Hangzhou, China}\\
}

\begin{document}

\maketitle

\begin{abstract}
Photoplethysmogram (PPG) and electrocardiogram (ECG) are commonly recorded in intesive care unit (ICU) and operating room (OR). However, the high incidence of poor, incomplete, and inconsistent signal quality, can lead to false alarms or diagnostic inaccuracies. The methods explored so far suffer from limited generalizability, reliance on extensive labeled data, and poor cross-task transferability. To overcome these challenges, we introduce QualityFM, a novel multimodal foundation model for these physiological signals, designed to acquire a general-purpose understanding of signal quality. Our model is pre-trained on an large-scale dataset comprising over 21 million 30-second waveforms and 179,757 hours of data. Our approach involves a dual-track architecture that processes paired physiological signals of differing quality, leveraging a self-distillation strategy where an encoder for high-quality signals is used to guide the training of an encoder for low-quality signals. To efficiently handle long sequential signals and capture essential local quasi-periodic patterns, we integrate a windowed sparse attention mechanism within our Transformer-based model. Furthermore, a composite loss function, which combines direct distillation loss on encoder outputs with indirect reconstruction loss based on power and phase spectra, ensures the preservation of frequency-domain characteristics of the signals. We pre-train three models with varying parameter counts (9.6 M to 319 M) and demonstrate their efficacy and practical value through transfer learning on three distinct clinical tasks: false alarm of ventricular tachycardia detection, the identification of atrial fibrillation and the estimation of arterial blood pressure (ABP) from PPG and ECG signals.
\end{abstract}

\section{Introduction}

Photoplethysmogram (PPG) and the electrocardiogram (ECG), have been extensively shown to be valuable for clinical diagnosis and prediction \cite{ppg_icu_review} \cite{ecg_icu_review}. However,in the context of clinical settings, such as Intensive Care Unit (ICU) or Operating Room (OR), the high incidence of poor, incomplete, and inconsistent signal quality can compromise the monitoring tasks, such as the prompt identification of life-threatening events. These physiological data are highly susceptible to factors such as patient movements, inadequate electrode-skin contact, as well as instrumental noise and artifacts \cite{quality_reason}. The consequences of such low-quality signals consist mainly in triggering false alarms that cause alarm fatigue of clinical staff \cite{FA}, affecting the accuracy of automated diagnostic systems \cite{ecg_quality_diagnosis} as well as compromising the continuous monitoring of these patients.

To solve this problem many methods have been proposed and they are mainly base on feature engineering coupled with traditional machine learning, such as length transforms \cite{zong2003robust} or peak energy detection \cite{oster2013open}, but the reliance on domain-specific prior knowledge restrict their generalizability. In response to these limitations, end-to-end deep learning models were introduced to automatically learn feature hierarchies and increase performance \cite{SQI_DL}. Recently, contrastive learning \cite{contrastive_False_alarm} and stable diffusion \cite{diffusion_False_alarm} have been proposed to address false arrhythmia alarms. However, the supervised learning paradigm requires vast quantities of meticulously labeled data to guide model training. The labeling process needs expert clinicians making it expensive and it is time-consuming. Furthermore, these models are typically designed for a single task. If we change the downstream task, the model needs to be redesigned and trained, demonstrating a fundamental lack of cross-task transferability (see Appendix A for further details on the state-of-the-art literature).

\begin{figure*}[t]
\centering
\includegraphics[width=1.0\textwidth]{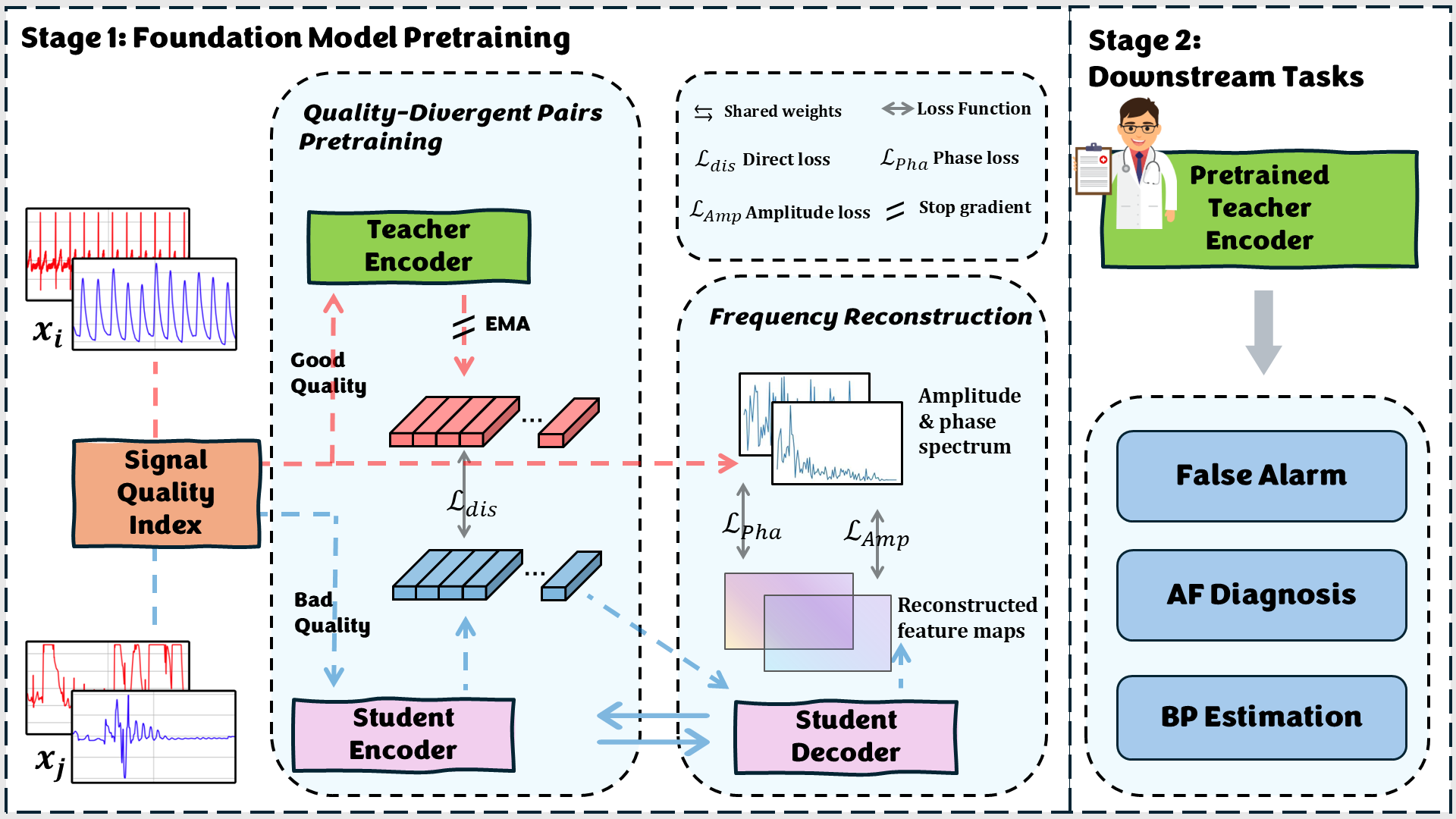} 
\caption{Pipeline of QualityFM pretraining and downstream tasks. \textbf{Stage 1 (Pretraining)}: a teacher encoder updated via Exponential Moving Average (EMA) guides a student encoder using pairs of high- and low-quality signals, while a decoder reconstructs spectral features. \textbf{Stage 2 (Downstream Tasks)}: the pretrained teacher encoder is adapted for downstream tasks, including false arrhythmia alarm detection, Atrial Fibrillation (AF) identification, and arterial blood pressure (BP) estimation.}
\label{overview}
\end{figure*}

To overcome the limitations of label dependency and poor transferability, we introduce a new paradigm: a multimodal physiological signal foundation model designed to build a general-purpose understanding of signal quality (qualityFM). The objective of this work is to construct a QualityFM from PPG and ECG recordings and to adapt it to various downstream tasks with the final goal to address challenges arising from poor signal quality and clinical needs. 
We pre-train our model using a tailored-made self-supervised learning strategy on a large-scale dataset consisting of 21 million waveforms of 30 second length for a total of 179,757 hours. The model input consist of paired physiological signals of differing quality close in time and a dual-track architecture was adopted to generate the corresponding feature representations, which serve for the pairs of different quality signals. We adopt a self-distillation strategy to use the encoder of high-quality signals as a teacher model to guide the encoder of low-quality signals. The teacher model parameters are smoothly updated using an Exponential Moving Average (EMA) strategy to ensure filter noises.

Furthermore, we employ a window-based Transformer to address the excessive computational complexity of self-attention in long sequential signals using a windowed sparse attention mechanism. This ensures computational feasibility while effectively modeling the essential local characteristics of the signals.

Finally, we design a Composite Loss Function with Direct and Indirect Supervision. A decoder, sharing weights with the encoder, aims at reconstructing the amplitude and phase of signal spectrum from the features. The loss is then calculated between the reconstructed spectra and the original spectra of the signal pairs.

To demonstrate its efficacy and practical value, we pre-train three models with varying parameter counts, from 9.6 M to 319 M, and validate their performance through transfer learning on three distinct ICU medical tasks. To the best of our knowledge, qualityFM represents the largest foundation model for addressing quality challenges of ECG and PPG signals in critically ill patients.

\section{Methodology}

\subsection{Datasets for Pretraining Task}
\subsubsection{VitalDB \cite{lee2022vitaldb}.} This database contains data collected from patients undergoing routine or emergency surgery in 10 of the 31 operating rooms at Seoul National University Hospital in South Korea. It includes a total of 6,388 surgical patients, encompassing 557,622 distinct data tracks, PPG and lead II ECG signals are sampled at 500Hz, they were successively down-sampled at 125Hz.

\subsubsection{MIMIC-III Waveform Database Matched Subset \cite{johnson2016mimic}.} It is a collection of waveform data obtained from bedside patient monitors in both adult and neonatal ICU. It comprises a total of 22,317 waveform records from 10,282 distinct ICU patients, sampled at a frequency of 125 Hz.

\subsection{Signal Quality Classification}
The description of the pre-processing analysis of PPG and lead II ECG signals can be found in Appendix B.
The PPG and ECG signals are firstly subdivided into overlapping segments, we use windows of length T=30 seconds, the signals are evenly sampled at $f_c=300$Hz, for a total of $N=9000$ samples for each signal segment.  $X_i \in \mathbb{R}^{C \times N}$ represent the combination of the two signal segments, where $C$ is the number of channels ($C=2$ in this case), and $t_i$ is the timestamp corresponding to the start of the segments.  Appendix C reports the details for the pipeline used to define the quality of the segments. In particular, the signal quality index for each segment $i$ is calculated as a weighted average of the signal quality index  of the two signal segments ($SQI_{ppg}$ and $SQI_{ecg}$) and a quality label $L_i$ is assigned to $X_i$. For the pre-training task, we construct segment pairs $X_i$,$X_j$, a pair is considered valid only if both segments are from the same subject, and their timestamps are within a 5-minute interval, and their signal quality labels are different. This selection criterion is formally defined as:
\begin{equation}
    |t_i - t_j| < \text{5 minutes} \quad \land \quad L_i \neq L_j
\label{positive pairs}
\end{equation}

\subsection{Self-Distillation based on Quality-Divergent Pairs}

Inspired by the self-distillation paradigms of DINO and iBOT architecture \cite{dino, ibot}, we employ a self-distillation framework tailored to capture robust physiological features from quality-divergent pairs. Given a pair as \{${(X_i, L_i), (X_j, L_j)}$\} where $L_i > L_j$, the framework assigns distinct roles to two parallel encoders. The high-quality segmnet $X_i$ is processed by a teacher encoder to obtain feature $U_t = E_{\theta_t}(X_i)$. Concurrently, the low-quality signal $X_j$ is processed by a student encoder yielding the representation $U_s = E_{\theta_s}(X_j)$, with ${\theta_t}$ and ${\theta_s}$ represent the parameters of the teacher and student encoders, respectively. The training objective is to align the output probability distributions of the student and teacher encoders. These distributions are generated by applying a softmax function to the encoder outputs. Specifically, for the student network, the probability distribution $P_s$ for a given input $X_j$ and feature $m$ is calculated as:

\begin{equation}
P_s(x)^{(m)}=\frac{\exp(E_{\theta_s}(X_j)^{(m)}/\tau_s)}{\sum_{m=1}^K\exp(E_{\theta_s}(X_j)^{(m)}/\tau_s)}
\label{KD}
\end{equation}

where $x\in X_j$, $\tau_s$ is a temperature parameter that controls the sharpness of the output distribution, $K$ is the dimension of the feature representation, and $m$ denotes the index of the feature. Similarly, $P_t(x)^{(m)}$ is computed from the teacher network's output. Then the student network updapte parameters by minimizing the cross-entropy loss between the two distributions:
\begin{equation}
\mathcal{L}_{\text{dis}} = - \sum_{(i, j)}\sum_{m=1}^{K} P_t(X_i)^{(m)} \log(P_s(X_j)^{(m)})
\label{eq:distillation_loss}
\end{equation}

The student's parameters $\theta_s$ are updated via backpropagation using stochastic gradient descent. In contrast, the teacher's parameters $\theta_t$ udapte based on an exponential moving average (EMA):
\begin{equation}
\theta_t \leftarrow \lambda \theta_t + (1 - \lambda) \theta_s
\label{eq:ema_update}
\end{equation}
where the momentum parameter $\lambda$  follows a cosine schedule, increasing from 0.996 to 1.0 throughout the training process. The EMA update allows the teacher model to evolve slowly and effectively filtering out noise. While the student model focuses on learning from local and individual sample information, the teacher model integrates knowledge across multiple samples and longer time horizons, thereby enhancing the model's generalization capabilities. The high-quality signal $X_i$ possesses a higher information density, and serving as the input to the teacher, it guides the student model to distill robust physiological patterns from the cluttered and low-quality signal $X_j$. The slow evolution of the teacher's parameters further refines this process by filtering out residual noise or artifacts even within the high-quality signals themselves. This synergistic and dual-track self-supervised learning approach thus confers the dual benefits of a smoothly filtered representation and superior generalization.

\subsection{Physiological Windowed Sparse Attention}
Standard Transformer-based models are often unsuitable for processing long physiological time series due to the self-attention mechanism's computational and memory complexity, which scales quadratically $(O(n^2))$ with the sequence length $n$ \cite{transformer}. To overcome this limitation, we adopt a sparse attention architecture. Considering the necessity of local context in extracting features from physiological signals, we specifically choose the sliding window attention mechanism for the encoders and decoders. This approach reduces the computational complexity to $O(n{\times}w)$, where $w$ is the fixed window size \cite{longformer}.

In the shallower layers of the windowed Transformer, the small receptive field of the windowed attention is adapt at capturing local waveform units, which is crucial for extracting the signal's morphological features. By stacking multiple layers of this windowed attention, the receptive field progressively expands in the deeper layers. This enables the model to learn the quasi-periodic characteristics over longer time spans \cite{guoremote}.

Furthermore, to enhance training stability, we apply Layer Normalization (LN) to the query (Q), key (K) and value (V) tensors before the matrix multiplication. It helps prevent excessively large values in the attention logits \cite{dehghani2023scaling}. The modified attention is calculated as follows:

\begin{equation}
\mathrm{Attention}(Q,K,V) = \mathrm{softmax}\left(\frac{\mathrm{LN}(Q)\mathrm{LN}(K)^T}{\sqrt{d_{\text{head}}}}\right)V,
\label{eq:ln_attention}
\end{equation}
where ${d_{\text{head}}}$ is the dimension of a single head in the multi-head attention mechanism.

\subsection{Direct and Indirect Loss Function}
The frequency-domain characteristics of physiological signals are related to their morphology and the waveforms that constitute them, e.g. the different P,Q,R,S,T waves on a single heart beat in the ECG signal. Therefore, we introduce an indirect loss term based on spectral reconstruction.

For any given high-quality input signal, $x_i(n)=\{x_i(0), x(1),\cdots,x_i(n)\}$, $\mathcal{P}=
\{0,1,\cdots,N\}$, we first apply the Discrete Fourier Transform (DFT):
\begin{equation}
X_i[k] = \sum_{n=0}^{N-1} x_i(n)\cdot e^{-j{2\pi}\frac{kn}{N}}
\label{eq:dft}
\end{equation}
then compute the ground-truth amplitude spectrum $A_i[k]$ and phase spectrum $\Phi_i[k]$ as follows:
\begin{equation}
A_i[k] = \sqrt{\mathrm{Re}(X_i[k])^2 + \mathrm{Im}(X_i[k])^2}
\label{eq:amplitude_spectrum}
\end{equation}
\begin{equation}
\Phi_i[k] = \mathrm{atan}(\mathrm{Im}(X_i[k]), \mathrm{Re}(X_i[k]))
\label{eq:phase_spectrum}
\end{equation}
where $\mathrm{Re}$ and $\mathrm{Im}$ denote the real and imaginary parts of a complex number, respectively. We then task a reconstruction head, implemented as a Feed-Forward Network (FFN), to predict these spectra from the student encoder's feature representation $U_s = E_{\theta_s}(X_j)$. Let the reconstructed amplitude and phase spectra be denoted as $\hat{A_j}$ and $\hat{{\phi}_j}$. We use the Mean Squared Error (MSE) loss to quantify the discrepancy between the reconstructed spectra and the spectra from the corresponding high-quality pair, $A_i$ and $\phi_i$, creating two reconstruction loss terms. The final integrated loss function is a weighted sum of the direct distillation loss and these two indirect reconstruction losses. The hyperparameters $\lambda_{Amp}$ and 
$\lambda_{Pha}$ control the relative importance of the amplitude and phase spectrum reconstruction tasks, respectively. 
Finally, the complete loss function in pre-training stage is:

\begin{equation*}\label{eq:complete loss}
\mathcal{L}_{\text{pre}} = \mathcal{L}_{\text{dis}} + \lambda_{Amp}\sum_{(i, j)\in\mathcal{P}}\|\hat{A_j}-A_i\|_2^2 +\\
\end{equation*}
\begin{equation}
\lambda_{Pha}\sum_{(i, j)\in\mathcal{P}}\|\hat{{\phi}_j}-{\phi}_i\|_2^2
\end{equation}

\section{Results}

\begin{table*}[htpb]
\centering
\caption{Performances values from different models on VTaC and MIMIC PERform AF databases.}
\label{table:performance_comparison}
\begin{tabular}{c | l | S[table-format=1.4] S[table-format=1.4] S[table-format=1.4] S[table-format=1.4] S[table-format=1.4] S[table-format=1.4]}
\toprule

\textbf{Database} & \textbf{Methods} & {\textbf{Acc}} & {\textbf{TPR}} & {\textbf{TNR}} & {\textbf{PPV}} & {\textbf{F1-Score}} & {\textbf{AUC}} \\
\cmidrule{1-1} \cmidrule{2-8}

\multirow{8}{*}{VTaC}
& ResNet-18 \cite{he2016deep} & 0.6369 & 0.7129 & 0.6089 & 0.4055 & 0.5161 & 0.9097 \\
& Transformer \cite{transformer} & 0.6674 & 0.7228 & 0.6477 & 0.4338 & 0.5415 & 0.8180 \\
& BeatGAN \cite{zhou2019beatgan} & {0.6413} & {0.1289} &{0.8328}  &{0.2224}  &{0.1621}  &{0.4978}  \\
& TAnoGAN \cite{bashar2020tanogan} & {0.6521} & {0.4162} &{0.7403}  &{0.3747}  &{0.3941}  &{0.7122}  \\
& CNN-CL \cite{contrastive_False_alarm}  & 0.7432 & 0.9075 & 0.6830 & 0.5118 & 0.6540 & 0.8835 \\
& Diffusion+CL \cite{diffusion_False_alarm}  & 0.6391 & 0.3521 & 0.7462 & 0.3411 & 0.3462 & 0.7274 \\
& SiamQuality \cite{ding2024siamquality} & {0.7268} & {0.6864} & {0.7420} & {0.5094} & {0.5848} & {0.7897} \\
\cmidrule{2-8}
& QualityFM-Base & 0.7767 & {0.7881} & {0.7722} & {0.5740} & {0.6642} & {0.8565} \\
& QualityFM-Large & \bfseries 0.8221 & \bfseries 0.7983 & \bfseries 0.8307 & \bfseries 0.6333 & \bfseries 0.7063 & \bfseries \textbf{0.9099} \\
& QualityFM-Huge & \textbf{0.8551} & \textbf{0.8220} & \textbf{0.8679} & \textbf{0.7080} & \textbf{0.7607} & 0.9047 \\

\cmidrule{1-1} \cmidrule{2-8}

\multirow{8}{*}{\makecell{MIMIC \\ PERform \\ AF}}
& BiLSTM \cite{huang2015bidirectional} & 0.5586 &  0.2994 & 0.6974  & 0.3435 & 0.3087 & 0.5466 \\
& Transformer \cite{transformer} & 0.6450 & 0.5592  &  0.6906 & 0.5044  & 0.5258 & 0.6704 \\
& CNN \cite{tang2020rethinking} & 0.8457 & \textbf{0.8341} & 0.8521  & 0.7521 &  0.7899 &  0.9199  \\
& MLP \cite{borghi2021atrial} & 0.6207 & 0.5028 & 0.6846 & 0.4691 & 0.4777 &  0.6463 \\
& SiamQuality \cite{ding2024siamquality} & {0.7321} & 0.7496 &  0.7226 & {0.5957} & 0.6579 & 0.8088 \\
\cmidrule{2-8}
& QualityFM-Base & \bfseries 0.8657 & \bfseries 0.8075 & \bfseries 0.8970   & \bfseries 0.8100 & \bfseries 0.8071 & \bfseries 0.9359 \\
& QualityFM-Large & {0.8671} & {0.8175} & {0.8937} & {0.8053} & {0.8102} & {0.9354} \\
& QualityFM-Huge & \textbf{0.8764} & {0.8299} & \textbf{0.9013} & \textbf{0.8199} & \textbf{0.8240} & \textbf{0.9411} \\

\bottomrule
\end{tabular}
\end{table*}

\subsection{Downstream Tasks}

\subsubsection{Identification of false ventricular tachycardia alarms.} A collection of physiological waveforms, named VTaC database \cite{lehman2023vtac}, contains labeled false ventricular tachycardia (VT) alarms from the ICU setting. The dataset comprises over 5,000 waveform records derived from 2,376 unique patients. Each record includes at least two electrocardiogram (ECG) leads along with one or more pulsatile waveforms, such as PPG and arterial blood pressure (ABP). Each waveform recording is sample at 125Hz and is a six-minute segment that captures the five minutes of physiological data preceding the onset of the VT alarm and the one minute following it. We select lead II ECG and PPG signals, then uniformly resampled at a frequency of 300 Hz. 

\subsubsection{Identification of atrial fibrillation.} The MIMIC PERform AF database was developed for the detection of atrial fibrillation (AF) using physiological signals \cite{bashar2019noise}. It contains physiological recordings of 35 critically ill adult patients, comprising 19 subjects with AF and 16 subjects with normal sinus rhythm. Each 20-minute recording consists of simultaneously acquired ECG, PPG and respiration signals, sampled at a frequency of 125 Hz. We select lead II ECG and PPG signals, then uniformly resampled at a frequency of 300 Hz. 

\subsubsection{Estimation of Arterial Blood Pressure values.} The UCI dataset is a widely-used public database and it is derived from the larger MIMIC-II Waveform Database \cite{kachuee2015cuff}. It comprises 11,844 records, corresponding to a total duration of approximately 719 hours, with all signals sampled at a frequency of 125 Hz. Each record contains synchronously captured PPG, ECG and arterial blood pressure (ABP) waveforms. For the ECG data, only the Lead II channel is provided.

\subsubsection{Training and Evaluation.} For the pre-training phase, we curated a large-scale dataset from two public sources, yielding 21,570,926 distinct 30-second samples for 179,757 hours in total. On this data, we designed and pre-trained three variants of our model: PSWA-Base (9.6M parameters), PSWA-Large (70M parameters), and PSWA-Huge (319M parameters). Unless otherwise specified, all experiments reported in this paper were conducted using the PSWA-Base model. For the evaluation of downstream tasks of VTaC, we followed the official splits for the training, validation, and test sets provided by the authors. For the MIMIC PERform AF and UCI databases, we partitioned the data into an 80\% training set and a 20\% test set, employing a cross-validation methodology to ensure the reliability of the results. All experiments were implemented using Python 3.11 and the PyTorch + CUDA , and executed on eight NVIDIA A100 80GB GPUs. The specific hyperparameters used in our training procedures are detailed in the Appendix D.

\begin{table*}[ht]
\centering
\caption{Performances value from the different models used to estimated blood pressure values, e.g. systolic blood pressure (SBP) and diastolic blood pressure (DBP), the less are the errors (MAE, ME, MASE), the better is the estimation.}
\label{tab:my_final_table}
\small
\setlength{\tabcolsep}{5pt}
\begin{tabular}{@{} l  S[table-format=2.2] c S[table-format=3.2]  S[table-format=1.2] c S[table-format=3.2] @{}}
\toprule
\multirow{2}{*} & \multicolumn{3}{c}{\textbf{SBP}} & \multicolumn{3}{c}{\textbf{DBP}} \\
\cmidrule(lr){2-4} \cmidrule(lr){5-7}
& {\textbf{MAE}} & {\textbf{ME $\pm$ SD}} & {\textbf{MASE (\%)}} & {\textbf{MAE}} & {\textbf{ME $\pm$ SD}} & {\textbf{MASE (\%)}} \\
\midrule

UNet \cite{ronneberger2015u} & 20.87 & {$ -3.31 \pm 26.05$} & 111.03 & 8.32 & {$ 1.65 \pm 10.85$} & 105.10 \\
ResNet \cite{he2016deep} & 22.13 & \textbf{0.05 $\pm$ 27.72} & 117.72 & 9.24 & {$-1.55 \pm 12.07$} & 116.80 \\
Spectro \cite{slapnivcar2019blood} & 20.64 & {$ 4.08 \pm 25.61$} & 109.78 & 9.15 & {$-0.96 \pm 11.92$} & 115.62 \\
VNet \cite{hill2021imputation} &21.86 & {$ 0.08 \pm 27.31$} & 116.30 & 9.12 & {$ 0.24 \pm 12.11$} & 115.22 \\
PPGIABP \cite{aguirre2021blood} &22.87 & {$ -5.13 \pm 27.82$} & 121.66 & 9.09 & {$ -0.695 \pm 11.97$} & 114.88 \\
SiamQuality \cite{ding2024siamquality} & 21.33 & {$ -0.91 \pm 26.84$} & 113.46 & 9.85 & {$ -1.82 \pm 12.55$} & 124.52\\

\cmidrule{1-7}
QualityFM-Base & 19.97 & {$ 5.54 \pm 23.84$} & 106.23 & 8.26 & {$ -1.25 \pm 10.36$} & 104.39 \\
QualityFM-Large & 19.46 & {$ 4.85 \pm 23.84$} & 103.51 & 7.86 & \textbf{0.11 $\pm$ 10.42} & 99.29 \\
QualityFM-Large & \textbf{19.32} & {$ 2.59 \pm 23.84$} & \textbf{102.93} & \textbf{7.61} & {$ -2.51 \pm 10.36$} & \textbf{98.81} \\

\bottomrule
\end{tabular}
\end{table*}

\subsubsection{Evaluation Metrics.} To comprehensively evaluate the performance of our models, we employed a standard set of statistical metrics. The following indicators were used to for classification tasks including VTaC and PERform AF databases: \textit{Accuracy (Acc)}: the overall proportion of correct classifications among the total number of samples; \textit{True Positive Rate (TPR)}: the proportion of actual positive cases that were correctly identified by the model; \textit{True Negative Rate (TNR)}: the proportion of actual negative cases that were correctly identified; \textit{Positive Predictive Value (PPV)}: the proportion of positive predictions made by the model that were actually correct; \textit{F1-Score}: the harmonic mean of Precision (PPV) and Recall (TPR), $2\frac{PPV\cdot \text{Recall}}{PPV+\text{Recall}}$; \textit{AUC}: the area under the ROC curve.

For the regression task of BP estimation on the UCI database, we used the following metrics: \textit{Mean Absolute Error (MAE)}: the average absolute difference between the predicted and actual values; \textit{(ME ± SD)}, Mean Error $\pm$ Standard Deviation: the ME quantifies the average prediction bias, while the SD measures the consistency and spread of the prediction errors; \textit {Mean Absolute Scaled Error (MASE)}: a scale-independent metric that compares the model's absolute error against the error of a naive baseline forecast, a value below 1 signifies that the model is more accurate than the naive prediction.

\subsubsection{Comparison with the State-of-the-Art performances.} Table~\ref{table:performance_comparison} provides the performance values of our proposed models against the state-of-the-art (SOTA) models. For the VTaC dataset, we followed the original benchmark protocol but utilized a distinct set of training parameters and settings (see Appendix D). QualityFM-Base consistently outperforms the existing SOTA approaches on both datasets. As the model size increases, from QualityFM-Base to QualityFM-Huge, we observe a consistent and significant improvement in all the performance metrics. This trend validates the superiority of our foundation model architecture and suggests that with access to even larger-scale datasets and by further increasing the model's parameter number, QualityFM can learn more comprehensive and deeper representations of physiological signals, leading to performance improvement on downstream tasks.

Table 2 presents shows the metrci used to compare the value predicted by our model and the SOTA models. We utilized the benchmark established by \cite{gonzalez2023benchmark}. The results demonstrate that our proposed approach achieves superior performance compared to the established methods on this dataset. It is evident that as the model's parameter number increases, the accuracy and reliability of the estimation improve accordingly.

\begin{figure}[ht]
    \centering
    \includegraphics[width=0.45\textwidth]{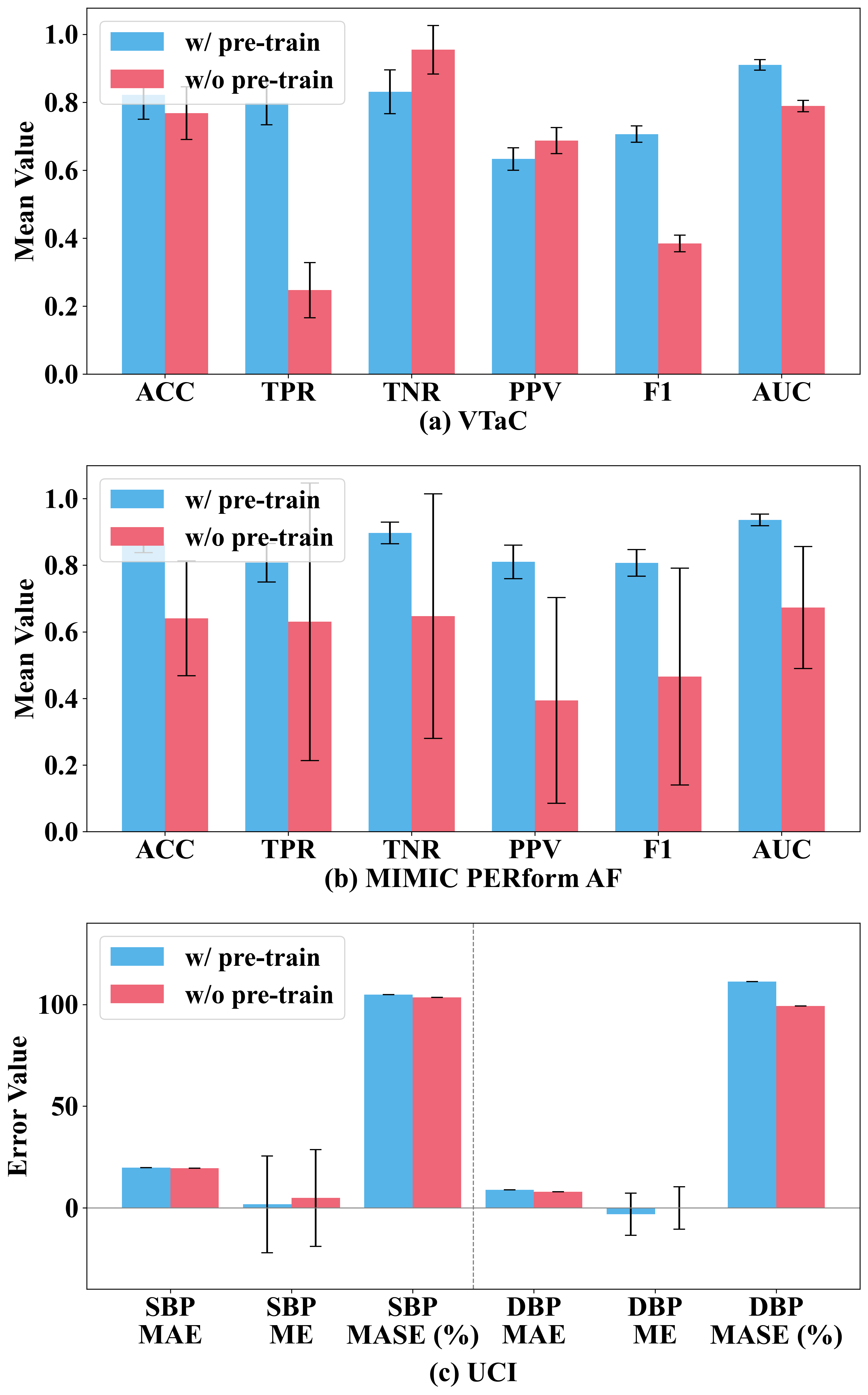}
    \caption{Comparison of the model performances with and without pre-training.}
    \label{fig:pretraining}
\end{figure}

\subsubsection{QualityFM Without Pre-training.}
In this experiment, we evaluated the impact of our pre-training strategy by training the QualityFM-Large model entirely from scratch, without initializing it with the pre-trained weights. Figure \ref{fig:pretraining} presents the performance of our model with and without pre-training for the three different downstream tasks. It is clear that the pre-trained model achieves significantly superior performance, with the advantage being most pronounced on the VTaC and MIMIC PERform AF datasets, i.e. classification taks. The pre-trained model's results on the VTaC dataset are particularly outstanding for the TNR and F1-score metrics, which indicates that the model can accurately identify False Alarms and thus help mitigate alarm fatigue. The performance on the UCI dataset is more comparable.

\subsubsection{Ablation Experiments on Windowed Sparse Attention.}
\begin{figure}[ht]
    \centering
    \includegraphics[width=0.45\textwidth]{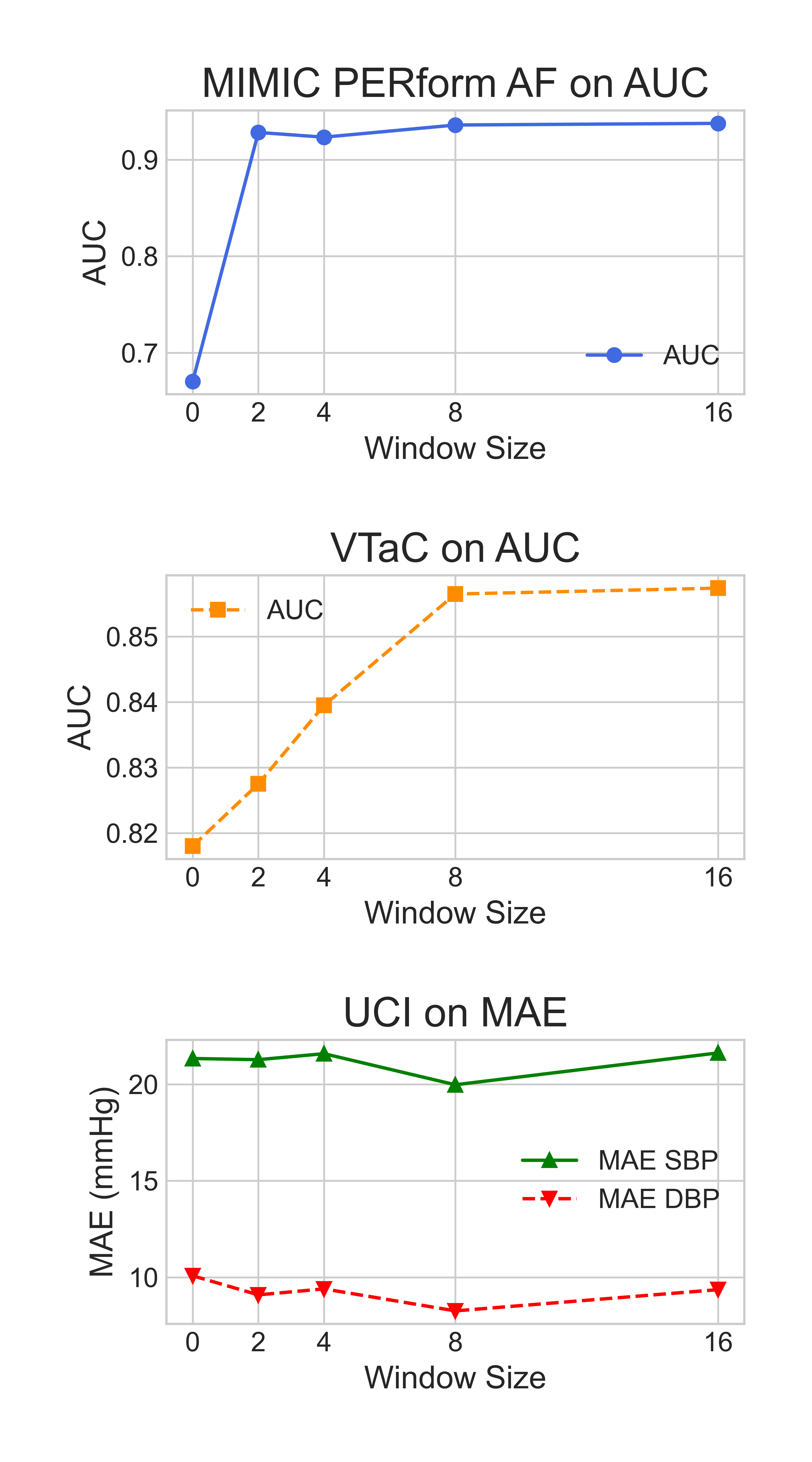}
    \caption{Comparison of the model performances with different window sizes of sparse attention.}
    \label{fig:window}
\end{figure}

To validate the effectiveness of our Windowed Sparse Attention mechanism and analyze the impact of its window size, we conducted an ablation study with varying window sizes: 0, 2, 4, 8, and 16. Figure \ref{fig:window} illustrates the results across three datasets. It is evident that the configuration with a window size of 0, which effectively deactivates the mechanism, yields the poorest performance. This finding confirms that Windowed Sparse Attention contributes positively to the modeling of physiological signals. Furthermore, as the window size increases, the performance gains begin to plateau and eventually degrade. Based on this trend, we selected a window size of 8 as the optimal configuration for our final model.

\subsubsection{Ablation Experiments on Backbone.}

\begin{figure}[ht]
    \centering
    \includegraphics[width=0.48\textwidth]{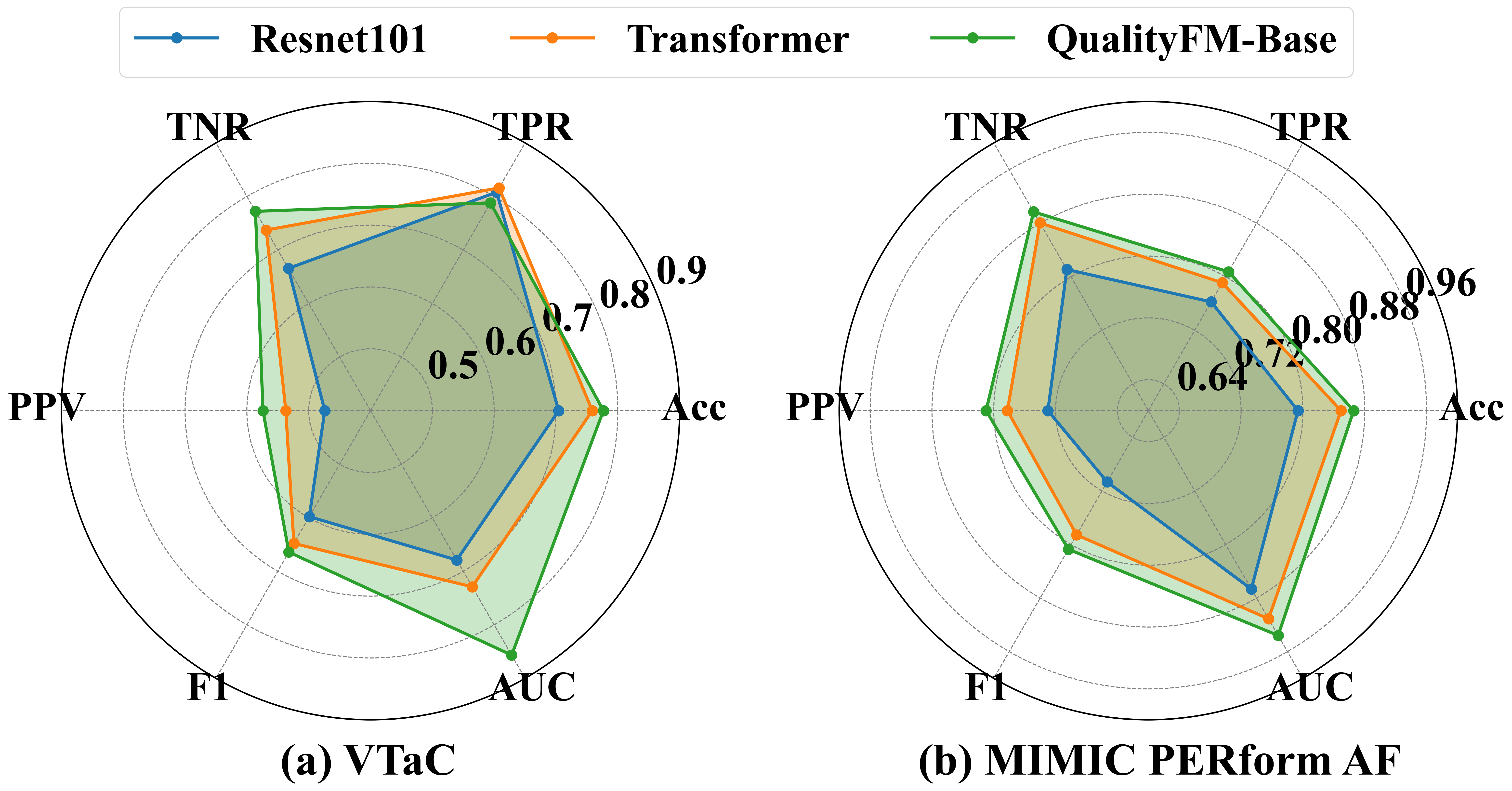}
    \caption{Comparison of QualityFM with different backbones architectures.}
    \label{fig:radar}
\end{figure}

To validate the superiority of our proposed feature extraction approach, we conducted an experiment comparing its performance against alternative backbone architectures. Figure \ref{fig:radar} displays this comparison using radar charts, which map the performance metrics on the VTaC and MIMIC PERform AF datasets for models using ResNet101, a standard Transformer, and our proposed QualityFM-Base. The results clearly indicate that the QualityFM-Base achieves a superior comprehensive performance profile compared to both the ResNet101 and standard Transformer architectures.

\subsubsection{Ablation Experiments on Indirect Loss.}
An ablation study was designed to validate the individual contributions of the amplitude and phase reconstruction losses within our proposed indirect loss framework. Table \ref{tab:loss on auc} presents the results of this study on the MIMIC PERform AF and VTaC datasets, evaluated using the AUC metric. For the regression task, Table \ref{tab:mae_performance} shows the corresponding outcomes on the UCI dataset, measured by the MAE for both SBP and DBP. The results consistently show that the introduction of either the amplitude reconstruction loss or the phase reconstruction loss yields a positive contribution to the model's performance. This finding underscores the importance of explicitly preserving frequency-domain information during the feature extraction process for physiological signals.

\begin{table}[h!]
\centering
\caption{Comparison of AUC scores for different combinations of indirect losses on the MIMIC PERform AF (AF) and VTaC databases.}
\label{tab:loss on auc}
\begin{tabularx}{0.45\textwidth}{@{} >{\centering\arraybackslash}X >{\centering\arraybackslash}X >{\centering\arraybackslash}X >{\centering\arraybackslash}X @{}}
\toprule
\textbf{$\lambda_{Pha}$} & \textbf{$\lambda_{Amp}$} & \textbf{AF} & \textbf{Vtac} \\ \midrule
\texttimes & \texttimes & 0.9150 & 0.8219 \\
\checkmark & \texttimes & 0.9107 & 0.8299 \\
\texttimes & \checkmark & 0.9320 & 0.8501 \\
\checkmark & \checkmark & \textbf{0.9359} & \textbf{0.8565} \\ \bottomrule
\end{tabularx}
\end{table}

\begin{table}[h!]
\centering
\caption{Performance Comparison of MAE scores for SBP and DBP Estimation on the UCI database.}
\label{tab:mae_performance}
\begin{tabularx}{0.45\textwidth}{@{} >{\centering\arraybackslash}X >{\centering\arraybackslash}X >{\centering\arraybackslash}X >{\centering\arraybackslash}X @{}}
\toprule
\textbf{$\lambda_{Pha}$} & \textbf{$\lambda_{Amp}$} & \textbf{SBP} & \textbf{DBP} \\ \midrule
\texttimes & \texttimes & 21.88 & 9.15 \\
\checkmark & \texttimes & 21.58 & 9.42 \\
\texttimes & \checkmark & 22.32 & 9.00 \\
\checkmark & \checkmark & \textbf{19.97} & \textbf{8.26} \\ \bottomrule
\end{tabularx}
\end{table}

\section{Discussion}
Our work introduces QualityFM, a multimodal physiological signal foundation model that addresses the critical and persistent challenge of poor and inconsistent signal quality in clinical environments. By pre-training on a large-scale dataset of over 21 million paired ECG and PPG waveforms, QualityFM develops a general-purpose understanding of signal characteristics, demonstrating impressive transferability and performance values on a variety of downstream clinical tasks. The consistent performance improvement observed as model size increases from the 9.6M  to the 319M parameters supports our core hypothesis: a scalable architecture pre-trained on extensive data can learn robust representations to mitigate signal quality issues.

A central innovation in our methodology is the self-distillation strategy using quality-divergent signal pairs. A teacher encoder, processing high-quality signals, guides a student encoder to extract meaningful physiological patterns from corresponding low-quality signals. This approach, augmented by a composite loss function that preserves critical frequency-domain attributes, enables QualityFM to outperform existing state-of-the-art methods that are often limited by their reliance on extensive manual labeling and task-specific designs. The significance of this pre-training is highlighted in our ablation studies, where a QualityFM model trained from scratch exhibited a marked drop in performance on the VTaC dataset. This underlines the efficacy of our pre-training in learning representations that are resilient to noise and artifacts.

Our ablation studies further validate these architectural decisions. The investigation into the Windowed Sparse Attention mechanism revealed that its absence (a window size of 0) leads to the poorest performance. As the window size gradually increases, performance slowly improves until it plateaus or even declines, confirming its positive contribution. Based on the experiments, a window size of 8 was chosen as the baseline. This configuration is adept at capturing local waveform morphology in the initial layers while progressively expanding the receptive field in deeper layers to learn long-range, quasi-periodic patterns. Furthermore, the ablation study on our indirect loss function confirmed the value of preserving frequency-domain information. The inclusion of both amplitude and phase reconstruction losses consistently improved performance on both classification and regression tasks, with the combined loss yielding the best results. Finally, when compared against alternative backbones, QualityFM-Base demonstrated a superior comprehensive performance profile, highlighting the effectiveness of our specialized architecture over more generic approaches.

Despite these promising results, we recognize several limitations. First, while QualityFM is a substantial model for its specific domain, its parameter count remains modest in comparison to the billion-parameter models emerging in other fields like natural language processing. Second, the current dual-channel input design, though multimodal, lacks the flexibility to dynamically adapt to a varying number of physiological signals without architectural changes. Finally, the substantial computational resources required for full fine-tuning of the largest model could impede its practical adoption.

Our future research will proceed along three main trajectories to address these limitations. We plan to expand our pre-training dataset to enable a significant scaling of the model's parameters, aiming to build a more comprehensive and knowledgeable foundation model. To enhance flexibility, we will investigate adaptive input mechanisms capable of handling a variable number and type of signals, drawing inspiration from recent advances in Large Language Models. Lastly, to improve computational efficiency and usability, we will explore parameter-efficient fine-tuning (PEFT) techniques, such as Low-Rank Adaptation (LoRA), as an alternative to full model fine-tuning for downstream task specialization.

\section{Conclusion}
In this work, we addressed the problem of low physiological signal quality in challenging clinical contexts by proposing a novel multimodal physiological signal foundation model. We pre-trained our model QualityFM on a large scale, utilizing over 21 million PPG and ECG signal segments to build a general-purpose understanding of signal quality. Our results on different downstream tasks demonstrate the good efficacy and transferability of our approach. This work not only establishes the largest foundation model for signal quality to date but also presents a robust pathway to overcome the bottlenecks of data labeling and task-specific model design. 

\section{Appendix A: Related Work}
\subsection{Physiological Signal Quality and ICU challenges.} 
Several works have been published to tackle this problems. \citet{behar2013ecg} utilized a set of Signal Quality Index (SQI), such as kurtosis based on signal amplitude distributions. \citet{li2012signal} proposed a machine learning method based on multi-signal quality assessment and feature fusion. Their approach employed a genetic algorithm and a Relevance Vector Machine, integrating 114 features from ECG, ABP, and PPG signals to effectively reduce the rate of false arrhythmia alarms in the ICU. More recent studies have shifted towards deep learning methodologies. \citet{liu2021ecg} developed an innovative hybrid model for ECG quality assessment, featuring a dual-input CNN architecture. One pathway automatically learned deep spectral features from the signal's S-transform spectrogram (a 2D time-frequency representation), while the other processed handcrafted statistical features. \citet{mousavi2020single} introduced a temporal attention unit between the CNN and LSTM layers of their model. This mechanism assign higher weights to signal regions most indicative of a specific alarm. Addressing noisy signals directly, the SQUWA architecture proposed by \citet{yan2024squwa} is a signal-quality-aware DNN designed specifically for detecting Atrial Fibrillation (AF) from noisy PPG signals. The model integrates an attention mechanism that is explicitly trained to assign greater weights to PPG segments exhibiting higher signal quality. \citet{liu2024ecg} employed an Large Language Model (LLM) to address the challenges of ECG denoising and missing value imputation. Their method progressively generates the output from a known sequence, which allows for flexible-length prediction while minimizing the accumulation of errors. This approach was shown to outperform state-of-the-art methods on tasks related to the identification of various cardiovascular diseases.

\subsection{Physiological Signal Foundation Models.}
The recent attempts to develop foundation models for physiological signals have primarily followed two main paradigms: masked signal reconstruction and contrastive learning. Methods in the first category draw inspiration from successful models in natural language processing, like BERT and GPT.   \citet{zhang2022maefe} introduced Maefe, a family of masked autoencoders with a specialized masking strategy that operates independently on both the temporal and lead dimensions, enabling the model to simultaneously learn morphological patterns and capture cross-lead correlations. developed by \citet{chen2025gpt}, adapted the Generative Pre-trained Transformer (GPT) architecture for PPG signals, focusing on leveraging the continuous and sequential characteristics of the data for autoregressive prediction. The second category focuses on contrastive learning, which learns representations by constructing positive and negative sample pairs. The OpenECG  project provided a systematic evaluation and benchmark of various self-supervised learning (SSL) methods across multiple downstream tasks \cite{wan2025openecg}. ECG-FM architecture  pre-trained its model by combining a contrastive learning task with a continuous signal masking target, merging the strengths of both paradigms \cite{mckeen2024ecg}. \citet{ding2024siamquality} proposed SiamQuality, a contrastive learning framework that forms pairs from high-quality and low-quality signal segments to specifically address signal quality. However, this approach has two key limitations. First, its reliance on a CNN architecture inherently focuses on local receptive fields, which can limit its ability to model the long-range and quasi-periodic changes in physiological signals \cite{chen2024actnet}. Second, the symmetric and identical dual-stream network may inadvertently encourage the model to disregard the intrinsic differences between clean and noisy signals.

\section{Appendix B: Pre-processing analysis for PPG and ECG signals}
We selected PPG and ECG waveforms from the available database. Any
recordings with a duration of less than 5 minutes or a missing value rate greater than 20\% of the duration of the signal were discarded. These signals were then segmented using a 30-second 50\% overlapping  window.
Within each segment, missing values were imputed using linear interpolation. The segments were resampled at the same sampling frequency 300 Hz. Finally, min-max normalization was applied to each segment to scale its amplitude values to a uniform range.

\section{Appendix C: Pipeline for the estimation of segment Signal Quality Index (SQI)}

The quality score $SQI_i$ of the considered segment $X_i$ was obtained by combined the SQI of each signal, i.e. $SQI_{ecg}$ and $SQI_{ppg}$, values ranging between 0 and 1.

\begin{figure}[h]
    \centering
    \includegraphics[width=0.45\textwidth]{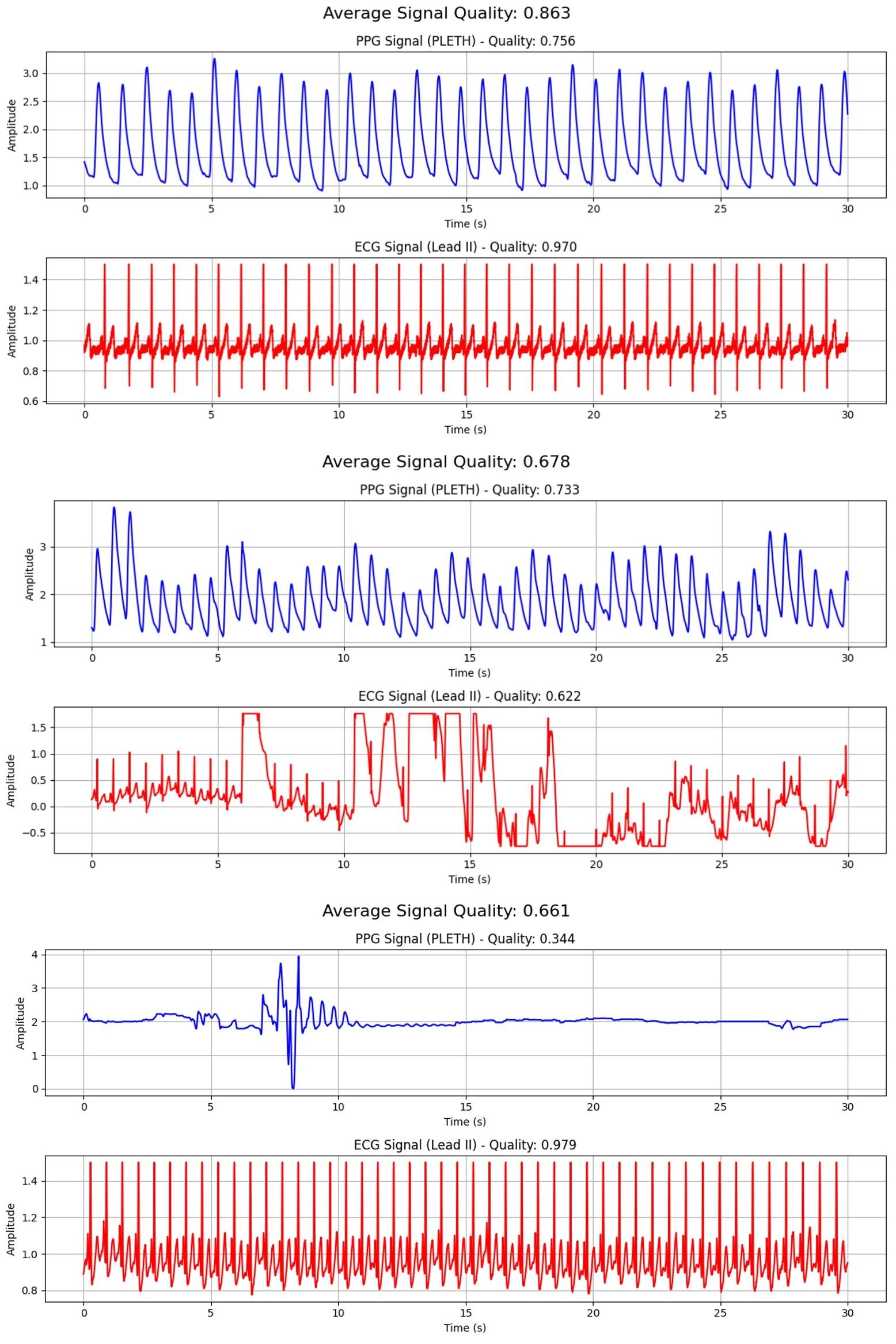}
    \caption{Example of ECG and PPG signals with different SQI values.}
    \label{fig:sqi}
\end{figure}

\textbf{PPG Signal Quality Index}. $SQI_{ppg}$ was obtained as a weighted average of these individual scores, commonly used in PPG analysis:
\begin{itemize}
    \item \textit{Signal Power Quality Index:} it reflects the signal-to-noise ratio of the waveform. It is typically calculated as the ratio of the power within the expected heart rate frequency band (e.g., 0.5-4 Hz) to the power in surrounding, noise-dominant frequency bands. A higher value indicates a cleaner signal.
    \item \textit{Perfusion Index:} it is the ratio of the pulsatile blood flow (AC component) to the non-pulsatile or static blood flow (DC component) in peripheral tissue. A higher PI generally indicates a stronger and more reliable signal.
    \item \textit{Skewness:} this is a statistical measure of the asymmetry of the signal's amplitude distribution. A clean PPG waveform exhibits a characteristic shape, and significant deviations from the expected skewness value can indicate the presence of motion artifacts or other noise.
    \item \textit{Relative Power:} given that the primary energy of the PPG signal is concentrated within a specific physiological frequency band (e.g., 1–2.25 Hz), this metric is calculated as the ratio of the Power Spectral Density (PSD) within this band to the PSD of the portion of spectrum mainly related to the signal (e.g., 0–8 Hz). A higher ratio suggests the signal is less contaminated by out-of-band noise.
    \item \textit{Entropy:} it provides a quantitative measure of the uncertainty or randomness present in the signal. A clean, periodic PPG signal has a predictable structure and thus lower entropy, while a signal corrupted by noise becomes more complex and unpredictable, resulting in a higher entropy value.
\end{itemize}

\textbf{ECG Signal Quality Index}. $SQI_{ecg}$ assesses signal quality combining two criteria: the quality of the recording and the proportion of identified heartbeats deemed to be reliable \cite{quality_reason}.
\textit{ECG Noise Quality Index.} It takes into account the signal-to-noise ratio of the raw ECG waveform used for beat detection. It is calculated by analyzing the ECG signal segments in order to assess the level of noise, it is based on the following indexes:
\begin{itemize}
\item \textit{Energy:} segments with abnormally high signal energy, which often indicates the presence of large-amplitude motion artifacts.
\item \textit{Sample Entropy:} segments with abnormally high signal complexity or erratic trends are considered irregular signal or contaminated by random noise.
\end{itemize}

\textit{Beat Signal Quality Index}. This index evaluates the physiological plausibility of the detected heartbeat sequence \cite{bashar2019noise, quality_reason}. Its calculation is based on the following criteria:
\begin{itemize}
\item \textit{Heart Rate Range:} It checks whether the instantaneous heart rate falls within a physiologically plausible range.
\item \textit{Rhythm Stability:} It assesses rhythm stability by detecting any sudden, excessive jumps between adjacent RR intervals or in the instantaneous heart rate. A lower proportion of such anomalous beats indicates a more stable rhythm.
\end{itemize}

Figure \ref{fig:sqi} shows some examples of signals and corresponding SQI values.

Finally, the SQI score is categorized into one of five discrete classes, which serve as pseudo-labels for our self-supervised learning framework. This classification is defined as follows:
\begin{equation}
\text{label} =
\begin{cases}
\text{"Excellent"} & \text{if } \text{SQI} \geq 0.9 \\
\text{"Good"} & \text{if } 0.7 \leq \text{SQI} < 0.9 \\
\text{"Acceptable"} & \text{if } 0.5 \leq \text{SQI} < 0.7 \\
\text{"Poor"} & \text{if } 0.3 \leq \text{SQI} < 0.5 \\
\text{"Bad"} & \text{if } \text{SQI} < 0.3
\end{cases}
\label{eq:quality_labels}
\end{equation}

\section{Appendix D: Hyperparameter Settings}
The following tables illustrate the hyperparameters used to develop our models.
\begin{table}[h]
    \centering
    \renewcommand{\arraystretch}{1.3}
    \captionsetup{skip=5pt}
    \caption{Hyperparameters for the pre-training.}
    \label{tab:hyperparameters_simple}
    \begin{tabular}{ll}
        \toprule
        \textbf{Hyperparameter} & \textbf{Value} \\
        \midrule
        Input length & 9000 \\
        Sampling rate (Hz) & 300 \\
        Batch size & 512 \\
        Sparse window size & 8 \\
        Learning rate & $1 \times 10^{-4}$ \\
        Learning rate scheduler & Cosine \\
        Optimizer & Adam \\
        Weight decay & 0.04 \\
        Total epochs & 10 \\
        Momentum & 0.7 \\
        Phase loss $\lambda_{Pha}$ & 0.5 \\
        Amplitude loss $\lambda_{Amp}$ & 0.5 \\

        \bottomrule
    \end{tabular}
\end{table}

\begin{table}[h]
    \centering
    \renewcommand{\arraystretch}{1.3}
    \captionsetup{skip=5pt}
    \caption{Hyperparameters used for the QualityFM models.}
    \label{tab:qualityfm_hyperparams_final}
    \begin{tabular}{l S[table-format=3.0] S[table-format=3.0] S[table-format=4.0]}
        \toprule
        \multirow{2}{*}{\textbf{Hyperparameter}} & \multicolumn{3}{c}{\textbf{QualityFM }} \\
        \cmidrule(lr){2-4}
         & {\textbf{Base}} & {\textbf{Large}} & {\textbf{Huge}} \\
        \midrule
        PWSA layers           & 2   & 21  & 50   \\
        Hidden size           & 512 & 512 & 512  \\
        MLP size              & 256 & 512 & 2048 \\
        Attention head number & 4   & 4   & 8    \\
        \bottomrule
    \end{tabular}
\end{table}

\begin{table}[th]
    \centering
    \renewcommand{\arraystretch}{1.3} 
    \captionsetup{skip=5pt} 
    \caption{Hyperparameters for downstream fine-tuning for teh ventricular tachycardia false alarm detection}
    \label{tab:finetuning_params}
    \begin{tabular}{ll}
        \toprule
        \textbf{Hyperparameter} & \textbf{Value} \\
        \midrule
        Input length & \num{9000} \\
        Sampling rate (Hz) & \num{300} \\
        Batch size & \num{512} \\
        Powerline frequency & \num{60} \\
        Learning rate & \num{1e-4} \\
        Optimizer & Adam \\
        Weight decay & \num{0.005} \\
        Total epochs & \num{500} \\
        Weighted class & \num{3.54} \\
        \bottomrule
    \end{tabular}
\end{table}

\bibliography{main}

\end{document}